\documentclass[paperpaper, 10 pt, conference]{ieeeconf}  

\IEEEoverridecommandlockouts                              

\usepackage{graphics} 
\usepackage{epsfig} 
\usepackage{mathptmx} 
\usepackage{times} 
\usepackage{amsmath} 
\usepackage{amssymb}  
\usepackage{amsmath, bm}
\usepackage{float}
\usepackage{subfigure}
\usepackage{threeparttable}
\usepackage{tabularx}
\usepackage{multirow}
\usepackage{booktabs}
\usepackage[linesnumbered,ruled,vlined]{algorithm2e}
\usepackage{algorithmicx}
\SetKwInput{KwInput}{Input}                
\SetKwInput{KwOutput}{Output}              

\title{\LARGE \bf
A Topological Solution of Entanglement for Complex-shaped Parts\\ in Robotic Bin-Picking
}

\author{Xinyi Zhang$^{1}$, Keisuke Koyama$^{1}$, Yukiyasu Domae$^{2}$, Weiwei Wan$^{1,2}$, and Kensuke Harada$^{1,2}$
\thanks{$^{1}$Graduate School of Engineering Science, Osaka University, Japan}%
\thanks{$^{2}$Artificial Intelligence Research Center, National Institute of Advanced Industrial Science and Technology (AIST), Japan}
}

\begin{document}

\maketitle
\thispagestyle{empty}
\pagestyle{empty}

\begin{abstract}
    This paper addresses the problem of picking up only one object at a time avoiding any entanglement in bin-picking. To cope with a difficult case where the complex-shaped objects are heavily entangled together, we propose a topology-based method that can generate non-tangle grasp positions on a single depth image. The core technique is the entanglement map, which is a feature map to measure the entanglement possibilities obtained from the input image. We use an entanglement map to select probable regions containing graspable objects. The optimum grasping pose is detected from the selected regions considering the collision between robot hand and objects. Experimental results show that our analytic method provides a more comprehensive and intuitive observation of the entanglement and exceeds previous learning-based work \cite{matsumura2019learning} in success rates. Especially, our topology-based method does not rely on any object models or time-consuming training process, so that it can be easily adapted to more complex bin-picking scenes.
\end{abstract}


\section{INTRODUCTION}\label{sec:intro}

    
    Bin-picking is commonly utilized in the industrial robot assembly line to provide and arrange necessary parts for assembly. It has been studied over the decades covering vision, planning, system integration, and solutions to specific tasks \cite{domae2014fast}, \cite{ghita2003bin}, \cite{kirkegaard2006bin}, \cite{oh2012stereo}, \cite{harada2016initial}, \cite{harada2018experiments}. However, there remain challenges if the shape of the target objects is complex. For instance, objects with S-shape or C-shape randomly placed in the bin are easily entangled with each other. In this case, the robot has difficulty in picking up only one object without any entanglement due to the complex physical collisions or the uncertainties in the environment. 
    
    A few studies address this issue in robotic bin-picking. Matsumura et.al \cite{matsumura2019learning} first tackle the problem of picking only one object from a stacked pile without causing entanglement \cite{matsumura2019learning}. A Convolutional Neural Network (CNN) is proposed to predict whether if the robot can pick up a single object among several pre-computed grasp candidates. They also use a physics simulator to collect training data by simulating bin-picking processes. However, we found some limitations in this research as follows. On one hand, data-driven method requires a large amount of training data and time-consuming training procedure. On the other hand, since CNN only makes predictions on cropped regions from the image, it cannot observe all the entangled objects from the whole input scene. Especially when the objects are heavily entangled, CNN may predict that all pre-computed grasp candidates share high possibilities of potential entanglement.
    
\begin{figure}[t] 
    \vspace{0.27cm}
    \centering
    \includegraphics[width=\linewidth]{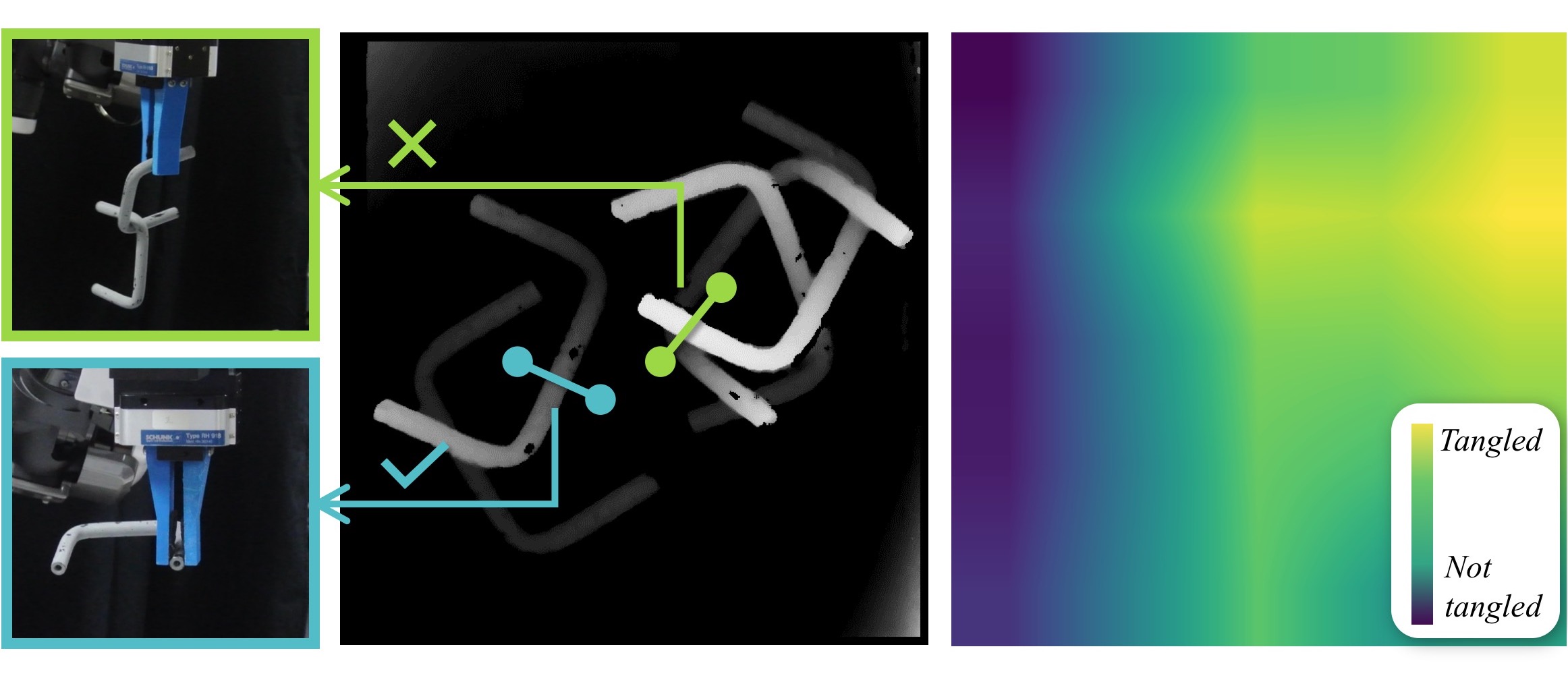} 
    \caption{Failure case and success case for a parallel jaw of picking up a single object from a random clutter using the proposed entanglement map. Grasp marked as green shows the failure example of picking up multiple objects. Grasp marked as blue is successfully generated avoiding the region containing tangled objects. }
    \label{fig:rq} 
    \vspace{-5pt}
\end{figure} 

    Motivated by the previous work, we propose an analytic approach to solve the entanglement issue in bin-picking using topological knowledge. In particular, we propose topology coordinates to obtain a series of metrics which can describe entanglement situation from a single depth image. Besides, we scan the input image in a sliding window manner to generate a feature map called entanglement map, which indicates the possibilities of containing entangled objects for each region. As Fig. \ref{fig:rq} shows, the entanglement map is able to discriminate which regions may contain tangled objects and which regions may not from a depth map. The regions marked as blue has high possibilities of containing graspable objects. Once the entanglement map is obtained, we select non-tangle regions and detect collision-free grasp candidates using graspability measure \cite{domae2014fast} respectively on selected regions. The output is a set of ranked grasp configurations of avoiding all entanglement and collisions. 
    
    Our main contributions are as follows. 
    
    1) We proposed a topology-based approach that can detect optimal grasps avoiding entanglement, which is a challenging problem in robotic bin-picking. 
    
    2) We fix the problems existing in previous work. Besides, our method only requires simple parameter tuning instead of time-consuming training and data collection. 
    
    3) We provide a complete observation and intuitive measurement of entanglement so that the bin-picking performance is improved dealing with complex-shaped parts. 
    
    4) We develop a vision-based bin-picking system and demonstrate a serious of experiments on a real robot. 
    
    Experiments suggest that our analytic method exceeds the previous learning-based method in success rates on a relatively difficult bin-picking task. 
    

\section{RELATED WORK}\label{sec:rel}
    
    Researches in grasp detection for bin-picking can be categorized into two approaches according to how the grasp is determined. Some researches focus on using shape or geometric information directly on a depth map or a point cloud \cite{domae2014fast}, \cite{ghita2003bin}, \cite{dupuis2008two}. Some researches use CAD models to better recognize the scene \cite{drost2010model}, \cite{choi2012voting}, \cite{chetverikov2002trimmed}. Recently learning-based method has been widely used to achieve better robustness and generalization \cite{matsumura2019learning}, \cite{mahler2017dex}, \cite{matsumura2018learning}. 
    
    However, robotic bin-picking faces some challenging tasks related to picking up complex-shaped objects. As described in Section \ref{sec:intro}, Matsumura et.al. propose a learning-based approach to train a classifier for several pre-computed grasp candidates from depth images and plan the grasp with the highest predicted score \cite{matsumura2019learning}. The major limitation of this work is the time-consuming training process and data collection. Besides, the network is only applied to the cropped regions of an input image, which may increase bias in the prediction. Other works such as \cite{leao2019perception} and \cite{leao2020detecting} also solve the problem by developing a geometric modeling method to fit multiple cylinders to an input point cloud and plan an singulation trajectory for bin-picking. Nevertheless, modeling and trajectory planning would be unstable when the shape of the object is complex. In this paper, we present an analytic approach without any object model or data collection, which makes it relatively easier to execute on a real robot. 
    
\begin{figure}[h] 
    \centering
    \includegraphics[width=0.65\linewidth]{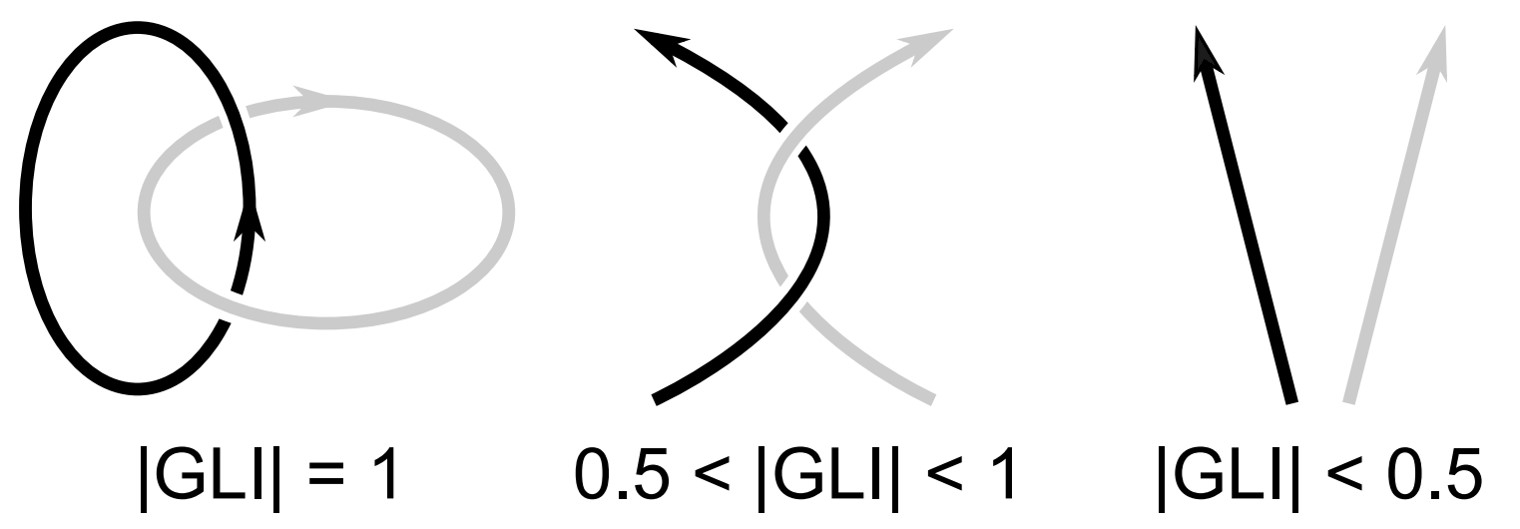} 
    \caption{GLI identifies whether two strands are tangled or not. }
    \label{fig:gli} 
\end{figure}

    Topological representation has been studied for decades and it is widely used for solving complex robotic motion synthesis. Ho and Komura firstly present the definition of topology coordinates to analyze the whole-body behaviors between two humanoid robots \cite{ho2007wrestle}, \cite{ho2009character}, \cite{ho2010controlling}. They calculate the topological relationship between two robot characters applying for different scenes. The most significant representation of entanglement would be Gaussian Link Integral (GLI) developed from knot theory \cite{reeke1988protein}. It describes a mathematical relationship between two tangled strands as Fig. \ref{fig:gli} shows. Moreover, topological representation plays an important role in robotic manipulation for deformable objects, such as tubes or ropes \cite{wakamatsu2004planning}, \cite{saha2007manipulation}, \cite{ivan2013topology}. This research is the first one to use topological knowledge in robotic bin-picking. The topological solution provides a more comprehensive measurements for dense clutters than the previous learning-based method. 
    

\section{Topology Coordinates for Bin-picking}\label{sec:topo}

    In this section, we introduce our revised topology coordinates based on the original theory proposed by Ho and Komura \cite{ho2009character}, \cite{ho2010controlling}. The details of the calculation are also presented. 

\subsection{Topology Coordinates}
    
    Topology coordinate is constructed between two tangled objects by three attributes \cite{ho2009character}, \cite{ho2010controlling}. The first attribute is \textbf{writhe}, which explains how much the two curves are twisting around each other. For instance, entangled objects get a higher score than separated objects. Writhe between two objects is calculated by Gaussian Link Integral (GLI) as follows. If we have two curves $\gamma_1$ and $\gamma_2$ which are point sets in Cartesian Coordinate, writhe can be calculated by GLI as follows.  

\begin{equation}
    GLI({\gamma}_1, {\gamma}_2) = \frac{1}{4\pi}\int_{{\gamma}_1}\int_{{\gamma}_2}\frac{d{\gamma}_1\times{d{\gamma}_2}\cdot({\gamma}_1-{\gamma}_2)}{||{\gamma}_1-{\gamma}_2||^3}
\end{equation}

    The second attribute is \textbf{density}, which describes how much the twisted area is concentrated on two curves. The third attribute is \textbf{center}, which is a location that explains the center location of the twisted area.

\subsection{Topology Coordinates in Depth Map}

\begin{figure}[h] 
    \centering
    \includegraphics[width=0.75\linewidth]{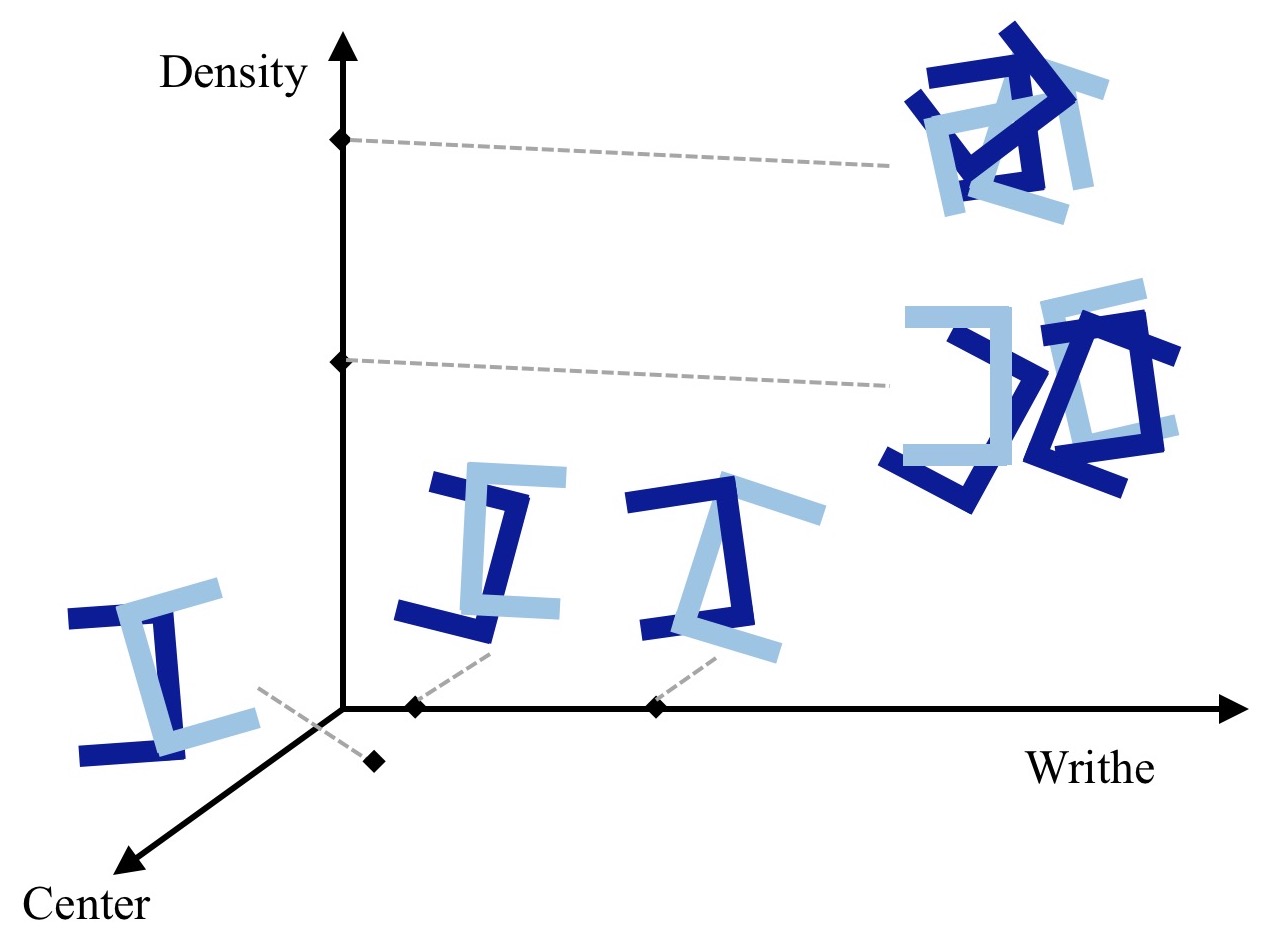} 
    \caption{Proposed topology coordinate illustrated by several rigid industrial parts. }
    \label{fig:topo-coor} 
\end{figure}

    Given a single depth image of a cluttered scene, the topology coordinate can be constructed to measure the entanglement (Fig. \ref{fig:topo-coor}). \textbf{Writhe} is a scalar attribute that indicates how much the objects are tangled together. A depth map containing tangled objects has higher writhe than the one with objects just overlapped together. \textbf{Density} is also a scalar attribute that indicates the distribution of the entanglement is evenly or intensively on the depth map. \textbf{Center} indicates the center position of entanglement on the depth map. 
    
    Original topology coordinates \cite{ho2009character} can only be applied for two characters and assume the exact position of characters are known. Different from the definition and calculation, we construct topology coordinates only using a single depth image containing multiple objects so that the position of each object remains unknown. Instead of computing the relationship between two objects, we extract the line segments of edge from the depth map and calculate the topology coordinates using the relationship between each pair of line segments. Edge contains all the information we need to describe the shape and position of objects. Even though the line segments of edges may not indicate the complete contours of all parts, the topological relationship calculated by line segments can still reflect how and where the entanglement occurs. 
 
\subsection{Calculation} \label{subsec:tc-cal}
    
    We use a depth map $I$ to compute topology coordinates $G=(w,\bm{c},d)$. In order to calculate these three attributes, we need to generate a matrix called writhe matrix $T$ firstly. Taken $I$ as input, we detect the edges and transfer them to a collection of 3-dimensional vectors $L=(l_1, l_2, ..., l_n)$. Writhe matrix $T$ is a $n\times{n}$ matrix that stores GLI of each segment pair in $L$. Particularly, instead of using Eq.(1), GLI between two 3-dimensional line segments is computed using the algorithm proposed by Klenin and Langowski \cite{klenin2000computation}. For instance, $T_{i,j}$ in the writhe matrix between i-th segment $l_i$ and j-th segment $l_j$ can be calculated by
\begin{equation}
    T_{i,j} = GLI(l_i,l_j)
\end{equation}

\begin{figure}[h] 
    \centering
    \includegraphics[width=\linewidth]{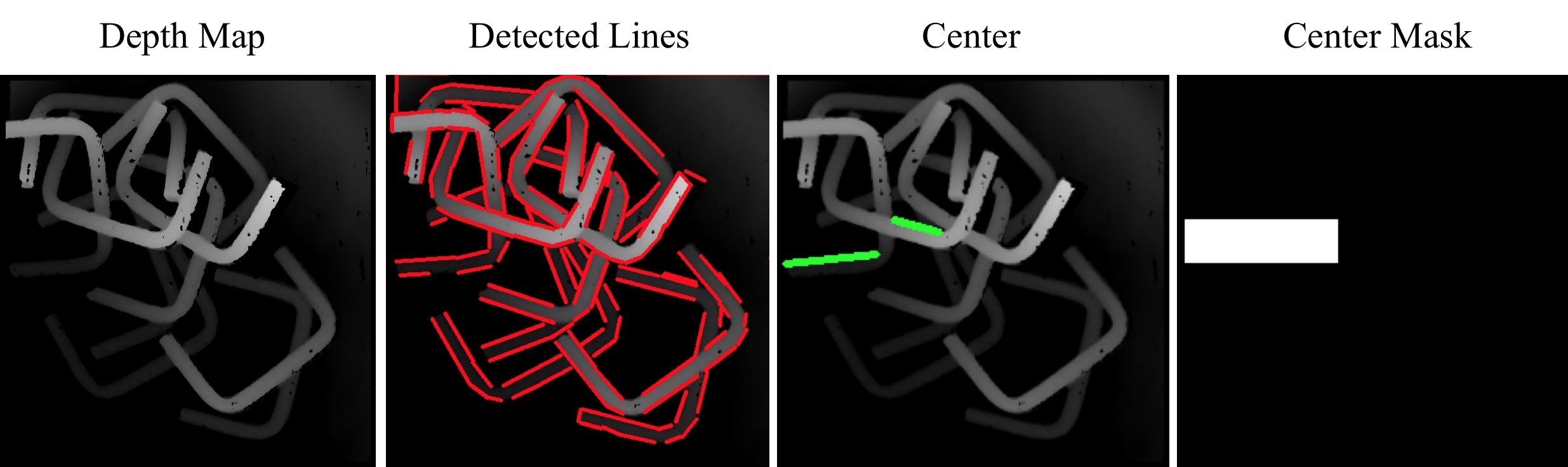} 
    \caption{Illustration of the center in a depth map. }
    \label{fig:coor-center} 
\end{figure}

    It can be seen that the writhe matrix $T$ is an upper-triangle-like matrix where half of the elements in $T$ are zero. We can compute the writhe $w$, density $d$ and center $\bm{c}$ using writhe matrix $T$. First, writhe $w$ is the sum of all values in $T$ divided by the number of line segments as follows. 
\begin{equation} 
    w = \frac{1}{n}{\sum_{i=1}^{n}\sum_{j=1}^{n}{T_{i,j}}}
\end{equation}
    Then, density $d$ is calculated by the ratio of the pairs that have higher values in writhe matrix $T$. We extract the non-zero elements from $T$ and compute $d$ using the number of elements higher than some threshold divided by the total number of non-zero elements. Here, we define the threshold as the mean of extracted non-zero elements. Finally, center $\bm{c}$ is simply obtained by the center of mass for matrix $T$, which is a segment pair that contributes the most to the entanglement. Moreover, we introduce a mask called center mask which has the same size as the input depth image (Fig. \ref{fig:coor-center}).  A center mask is a binary matrix with an area consisting of both center segments.

\subsection{Explanations}
    
    Let us explain how the topology coordinates service for the entanglement using a depth map. 
    
\begin{figure}[t] 
    \centering
    \vspace{2mm}
    \includegraphics[width=\linewidth]{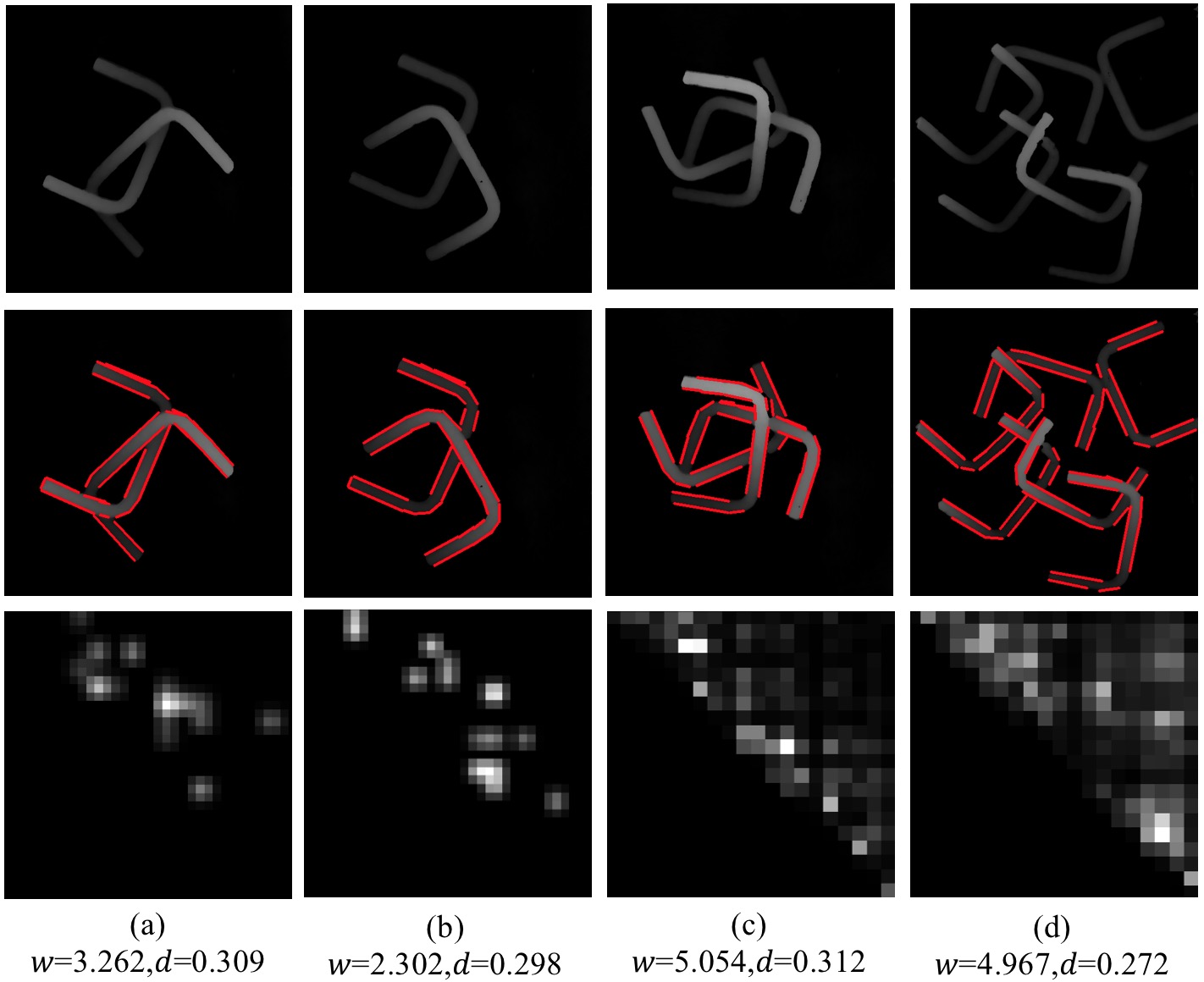} 
    \caption{Input depth images, detected edge segments, and visualized writhe matrices for 4 different clutter patterns. (a-b) Scenes with different writhe and similar density. Overlapped objects in (b) has lower writhe. Writhe value $w$ and density $d$ is also written. (c-d) Scenes with different density and similar writhe. Visualized writhe matrix show that for the sparse clutter such as (b), density would be lower. Writhe value $w$ and density $d$ is also written.}
    \label{fig:coor-cprs} 
\end{figure}

    Firstly, writhe is a quantitative measure that denotes how much the entanglement is on the depth image, while density and center denote the position information for the entanglement. Therefore, we present the visualization of the writhe matrix and the center mask to elaborate the density and center more intuitively. 
    Fig. \ref{fig:coor-center} presents which region is the entanglement center from the input depth image while Fig. \ref{fig:coor-cprs} shows four different clutters with various writhe and density values by presenting the corresponding input depth images, detected edge segments, and visualized writhe matrices. This visualized matrix derives from $T$ in Section.\ref{subsec:tc-cal} since it only remains the larger elements and is resized to a certain size. We can observe which pairs of edge segments share the larger writhe value and what is the distribution of the segment pairs from the matrix. If the edge segments are tangled with those near them, the brighter values are concentrated around the axis of the matrix. On the contrary, if the edge segments share rather larger writhe values with those all over the image, the distribution in the writhe matrix is rather even. Therefore, the more concentrated around the axis in the writhe matrix, the higher the corresponding density is. The details are elaborated as follows. 

\begin{figure*}[t]
    \centering
    \vspace{2mm}
    \subfigure[Entanglement map generation. ]
    {\includegraphics[width=0.5\linewidth]{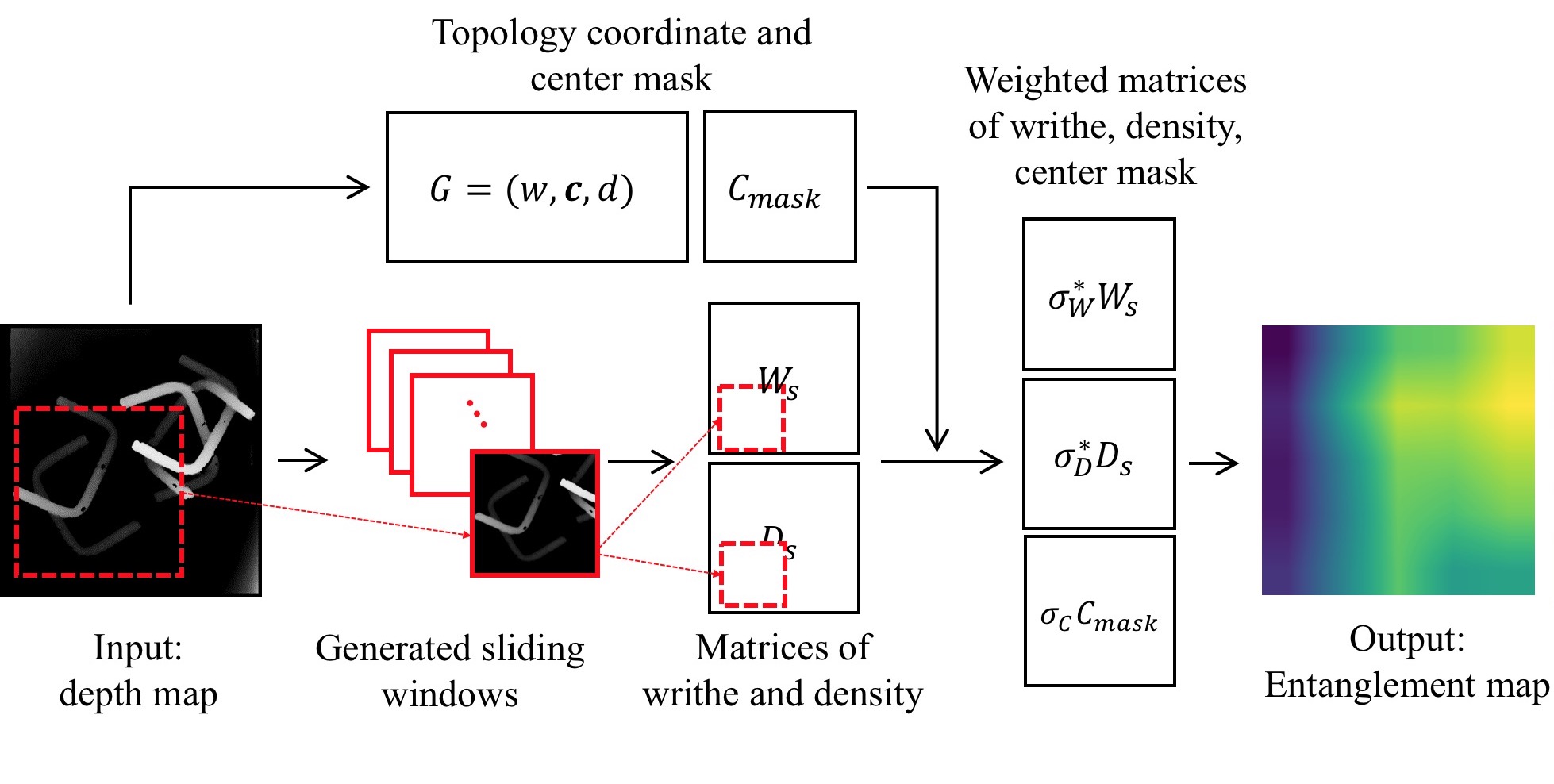}}
    \hspace{2mm}
    \subfigure[Grasp detection. ]
    {\includegraphics[width=0.42\linewidth]{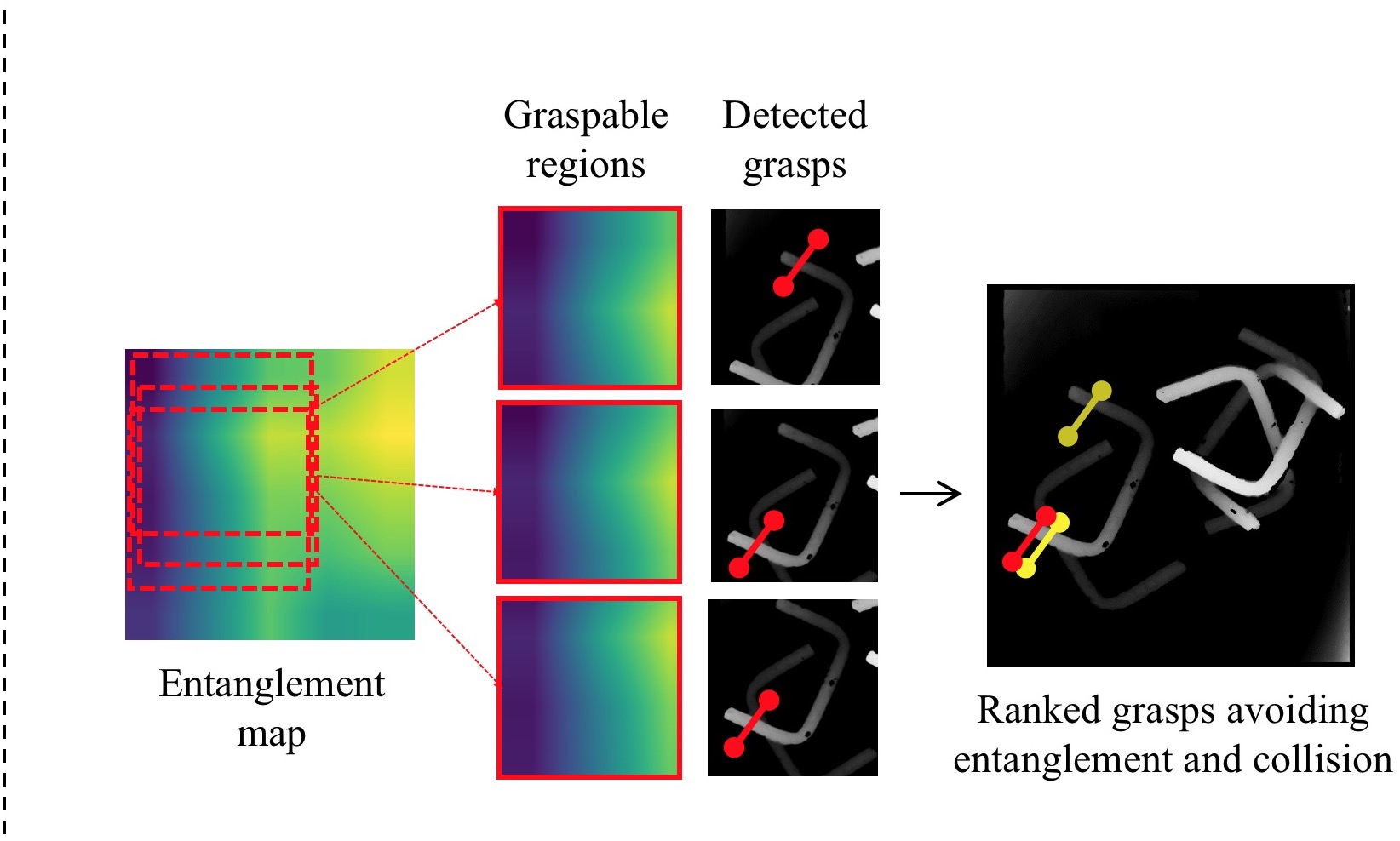}}
    \caption{Proposed grasp detection method to generate optimal grasps avoiding any entanglement and collision. }
    \label{fig:method}
\end{figure*} 

    \textbf{Writhe.} Fig. \ref{fig:coor-cprs}(a) is the situation where two objects are twisting together while Fig. \ref{fig:coor-cprs}(b) refers to two simply overlapped objects. They have similar density $d$ but differ from writhe $w$. The writhe of twisted objects is larger than overlapped ones. If the robot wants to pick objects from these scenes, Fig. \ref{fig:coor-cprs}(b) with lower writhe has s higher possibility of a successful picking. 
    
    \textbf{Density.} From Fig. \ref{fig:coor-cprs}(c-d) we can tell by the human observation that Fig. \ref{fig:coor-cprs}(d) would be a better choice for robot simply by taking a look. Visualized writhe matrix and density value can also explain the scene numerically. The visualized matrix in Fig. \ref{fig:coor-cprs}(d) has an even distribution of brighter pixels since every line segment tends to tangle with more segments. Because as the number of segment pairs with larger GLI increases, the number of bright pixels in the visualized matrix also increases. Thus, these brighter pixels distribute more evenly, in other words, the density becomes smaller. On the contrary, every object in Fig. \ref{fig:coor-cprs}(c) is twisted with the other objects, thus, the entanglement is distributed intensely on the depth map. For the visualized matrix, the pixels with larger writhe are concentrated around the axis. Therefore, when writhe values are similar, density can also contribute to entanglement analysis. 
    
    \textbf{Center.} The center is computed by the center of mass of the writhe matrix, which is a pair of line segments that contributes the most to the entanglement. We present how the center affects the entanglement by presenting a mask that has the same size as the input depth image (Fig. \ref{fig:coor-center}). The center mask indicates the position information of the entanglement but not as much as writhe and density do. 
    
    To summarize, by focusing on the metrics of the entanglement regardless of the number of objects, situations with lower writhe and lower density is preferred. Therefore, the topology coordinate proposed in this section can be used to determine where a non-tangle grasp should be located. 


\section{Grasp Detection of Avoiding Entanglement}\label{sec:grasp}

    This section elaborates grasp detection method for picking up only one object by measuring the entanglement metrics using the proposed topology coordinates. 

\begin{figure}[h] 
    \centering
    \includegraphics[width=0.9\linewidth]{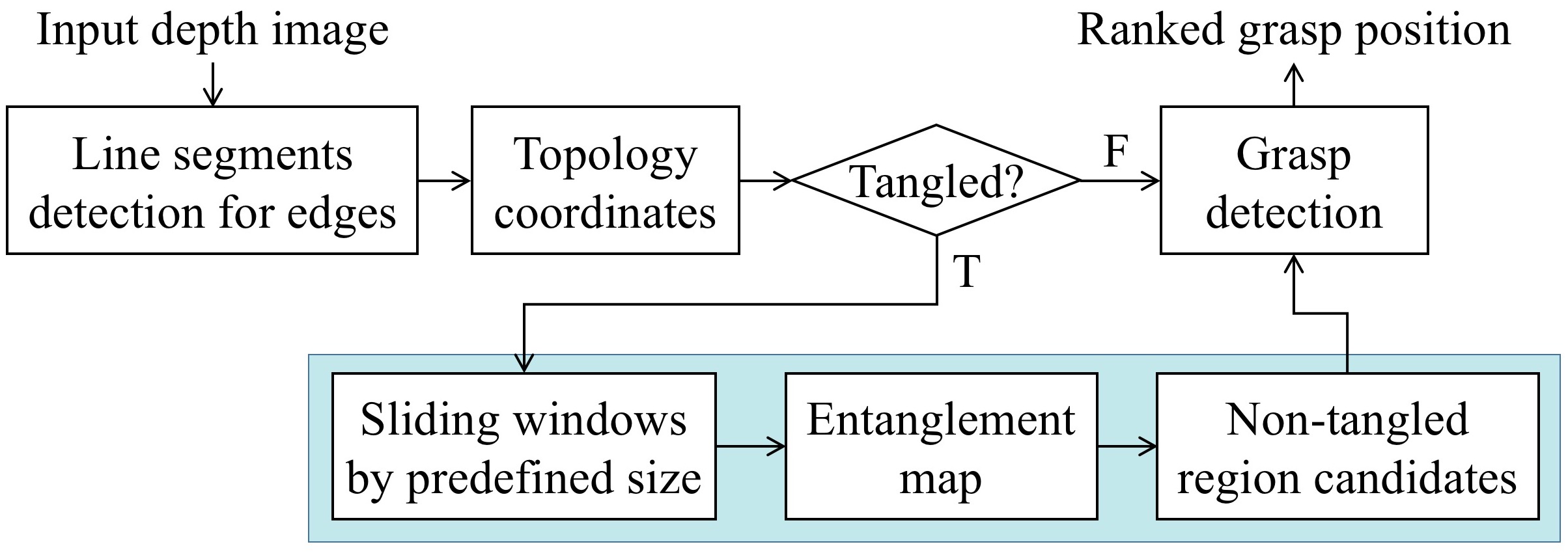}
    \caption{Proposed bin-picking pipeline. }
    \label{fig:overview} 
\end{figure}

\subsection{Overview}

    We select a parallel jaw gripper and use a single depth map as input to compute grasp hypotheses. The overview of the proposed grasp detection method is illustrated in Fig. \ref{fig:overview}. First, a depth map is captured and used to construct the topology coordinate. Then, we use the writhe in topology coordinates to determine if the objects in the bin are tangled. If not, grasp is detected only considering the collision. If the entanglement exists, we calculate the entanglement map which evaluates where the potential tangled parts are in the box. We crop several regions with high possibilities of containing graspable objects from the entanglement map. Finally, grasp is detected and ranked in each selected region using graspability measure \cite{domae2014fast}. 

\subsection{Entanglement Representation: Entanglement map}

    We explain how to compute an entanglement map by a given depth map $I$. First, we compute the line segments for edges on input depth map $I$, and generate topology coordinate $G=(w,\bm{c},d)$ along with center mask $C_{mask}$ for $I$ . Then, we use a pre-defined sliding window function for $I$ to obtain expanding information. For each window, we calculate its own writhe and density. This sliding window function returns two matrices $W_s$, $D_s$, which respectively store writhe and density of each region. The combination of two matrices refers to the rough entanglement information on each regions from $I$. However, we would like to precisely evaluate the entanglement situation upon the whole image. We use calculated topology coordinate $G$ to evaluate the weights for these matrices and center mask. The initial weights are manually defined as ${\sigma}_W=0.8$, ${\sigma}_D=0.15$, ${\sigma}_D=0.05$ respetively for $W_s$, $D_s$ and $C_{mask}$ since writhe affect more on predicting potential tangled regions. If $\overline{d}$, average of $D_s$, is larger than $d$ of the coordinate $G$, it means that density may affect the result of entanglement map generation. Therefore, the weights are modified as, 
\begin{equation}
    {\sigma}_{D}^*=(\overline{d}/d)\sigma_{D};\quad {\sigma}_{W}^*=1-{\sigma}_{D}^*-{\sigma}_C
\end{equation}
    The center mask is independent of the sliding window algorithm so that the weight ${\sigma}_C$ remains the same. Finally, entanglement map $E$ is obtained by the addition of weighted metrics as Eq.(\ref{eq:final-map}) shows following a bi-linear interpolation. 
    
\begin{equation}
    E={{{\sigma}_W}^*}W_s+{{\sigma}_{D}^*}D_s+{{\sigma}_{c}}C_{mask}
    \label{eq:final-map}
\end{equation}

\begin{figure}[t] 
    \centering
    \vspace{2mm}
    \includegraphics[width=0.9\linewidth]{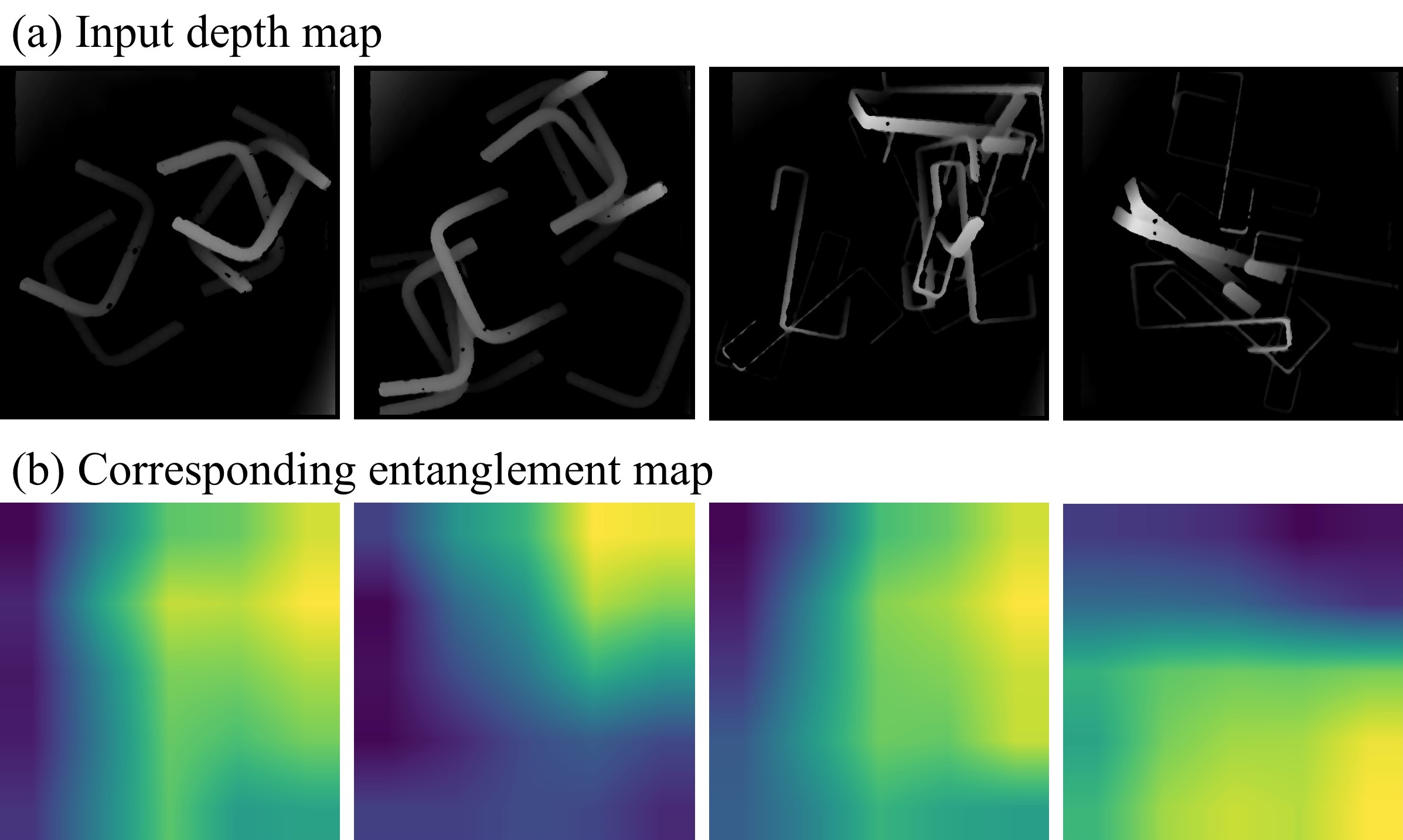} 
    \caption{Input depth map and corresponding entanglement map. Yellows stands for entangled region. Blue area is the region where has low possibility of entanglement. }
    \label{fig:tangle-map} 
\end{figure}

    Some examples are presented in Fig. \ref{fig:tangle-map}. In our perspective, the entanglement map is the visualization that indicates possibilities of entanglement in every region for the whole depth map. We can observe those areas where objects are heavily tangled with each other are marked as yellow, while blue areas refer to non-tangled regions. We prefer to generate grasps on blue areas. 
    
\subsection{Grasp Detection}

    We present a grasp detection method considering both entanglement and collisions. Another sliding window function is executed to seek several region candidates with smaller values from the entanglement map. In each region, a grasping pose can be computed using graspability measure \cite{domae2014fast}. Graspability is an index for detecting a grasping point by convoluting a template of contact areas and collision areas for a robot hand. To put it more precisely, it is based on the idea that the object should be in the trajectory of hand closing, and there should be no object in the position to lower the robot hand. We use a parallel jaw gripper and rotate the gripper template along for 4 orientations. For the detected grasp candidates in each region, we rank them simply by pixel values in the entanglement map of the corresponding positions combined with the graspability score. Finally, the best grasp position is selected as the top of ranked grasp candidates.


\section{Experiments and Results}\label{sec:exp}
\subsection{Experiment Setup}

\begin{figure}[h] 
    \centering
    \vspace{2mm}
    \includegraphics[width=\linewidth]{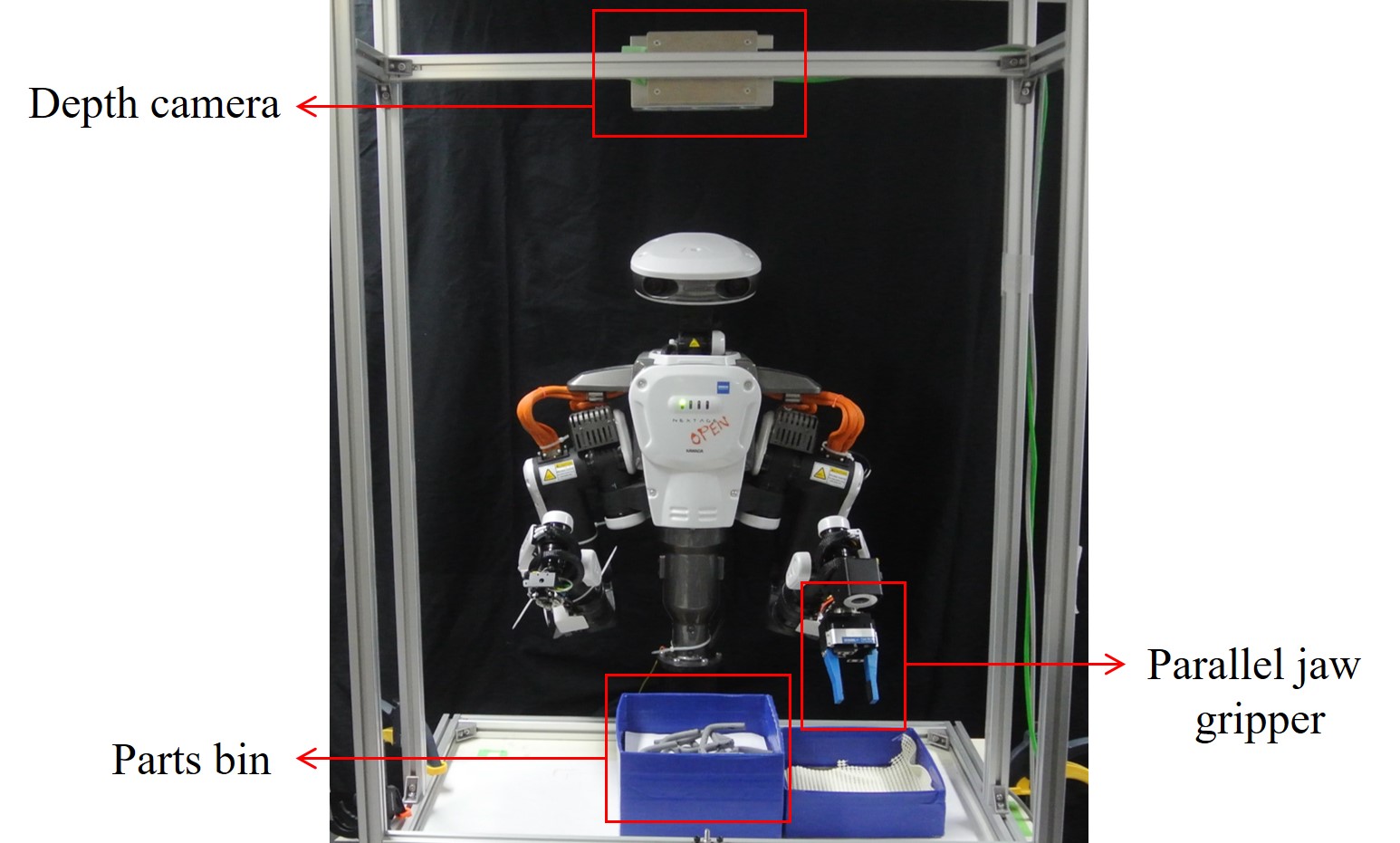} 
    \caption{Experiment setting. We fix a 3D camera at the height of 1 meter above the parts bin. We select the parallel jaw gripper to execute picking. }
    \label{fig:exp-system} 
\end{figure}

    We perform several real-world robot experiments to evaluate our method in bin-picking. We use NEXTAGE from Kawada Robotics and set a fixed 3D camera YCAM3D-II one meter straight above from the bin. We use Choreonoid and graspPlugin to simulate and execute the movement of the robot. The execution time was recorded on a PC running Ubuntu 16.04 with a 2.7 GHz Intel Core i5-6400 CPU. Our experiment system is set as Fig. \ref{fig:exp-system} shows. 

\begin{figure}[h] 
    \centering
    \includegraphics[width=0.85\linewidth]{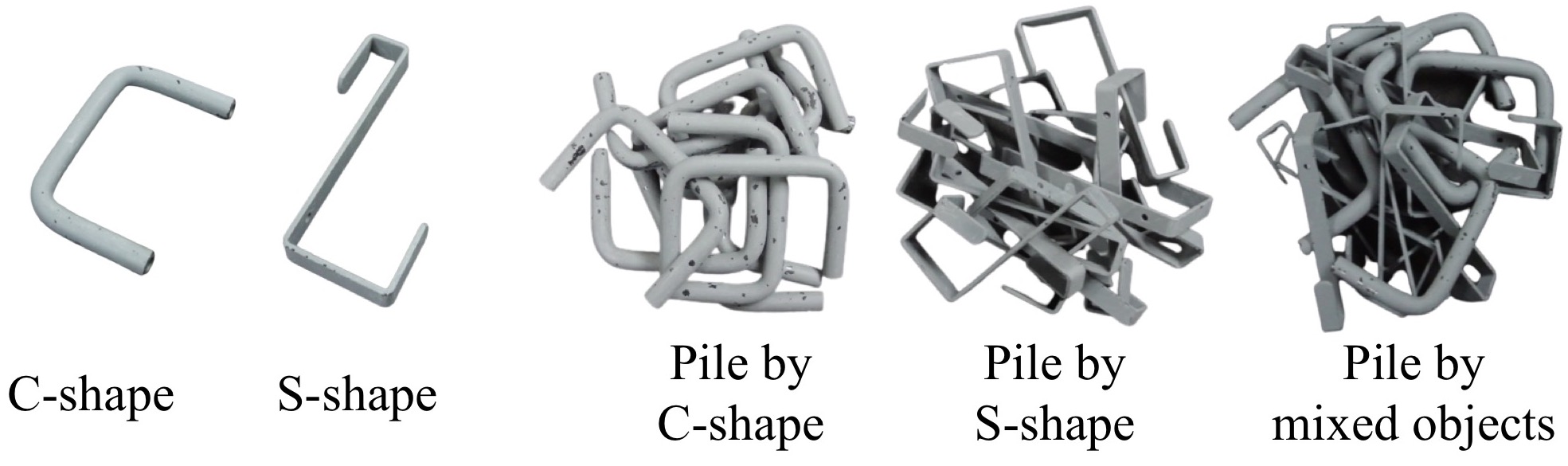} 
    \caption{Types of objects and pile patterns used in the experiment.  }
    \label{fig:exp-object} 
\end{figure}

\begin{figure*}[t] 
    \centering
    \vspace{2mm}
    \includegraphics[width=0.85\linewidth]{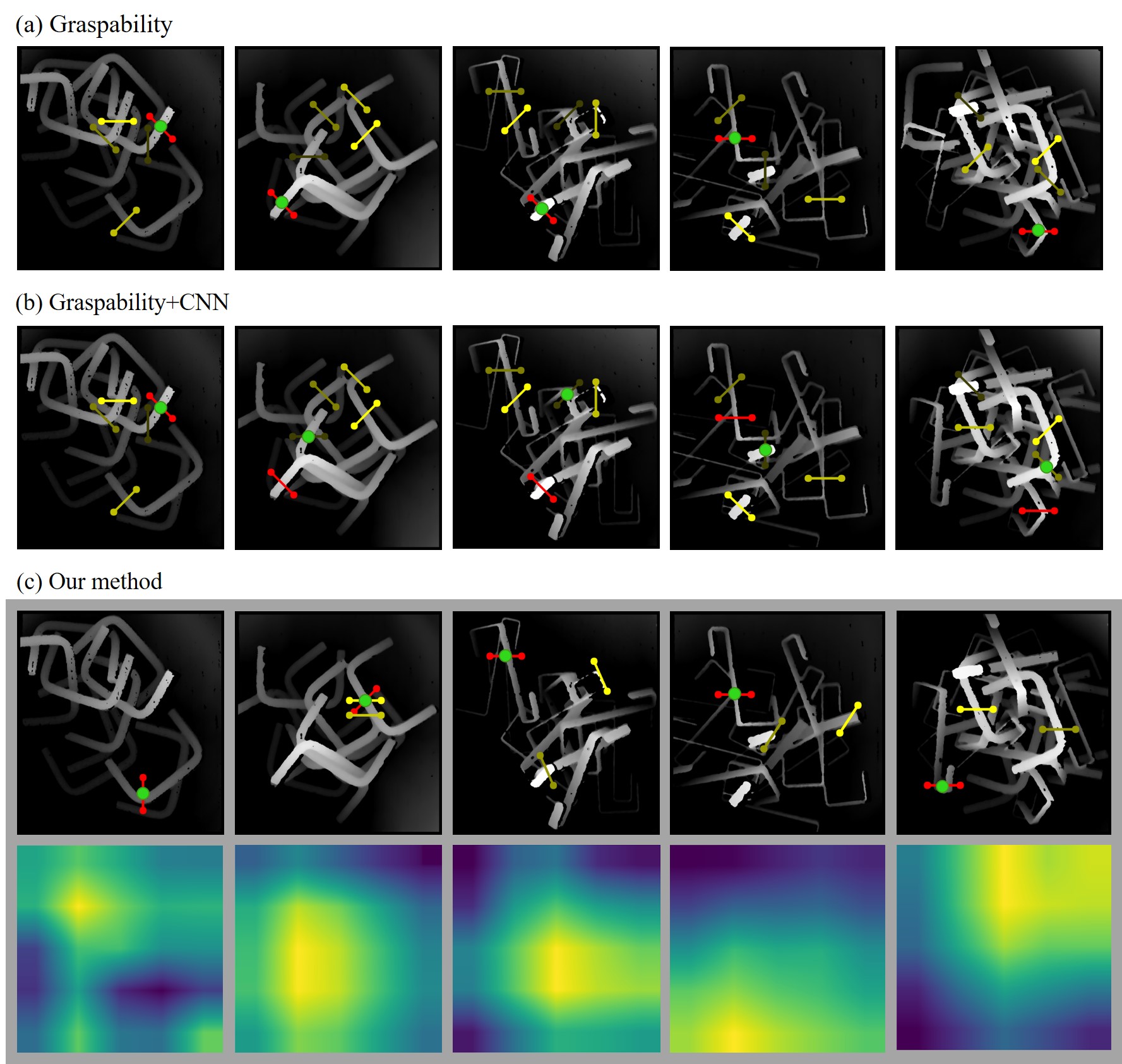} 
    \caption{Experiment results using the same depth maps. Best grasps are emphasized using green dot. (a) Results of \textit{Graspability}. Grasps marked as red denote to the best grasp. Yellow refers to grasps with the ranked order of 2nd, 3rd, 4th, and 5th. (b) Results of \textit{CNN}. The grasp candidates are the same as (a) while \textit{CNN} predicts the best grasps marked as green. (c) Result of proposed method. Both depth maps and entanglement maps are presented. Red denotes to the best grasp. Yellow refers to grasps with the ranked order of 2nd and 3rd. }
    \label{fig:res} 
\end{figure*}
    
    As Fig. \ref{fig:exp-object} shows, two types of industrial parts with complex shapes are selected. We prepare three patterns of clutter state by only C-shaped objects, only S-shaped objects, and mixed objects. In particular, three picking trials are performed for each clutter. Each trial contains 20 times of picking to record the success rate of only picking one object. 
    
    The purpose of the experiment is to compare our method with two baselines. 
    
    1) \textit{Graspability}. We select graspability measure \cite{domae2014fast} which is a general grasp detection algorithm only using a hand template. It outputs several grasp candidates ranked by graspability measure. 
    
    2) \textit{CNN}. We also select previous work from \cite{matsumura2019learning} which is the first approach to predict potential tangled objects in bin-picking. It takes the same grasp candidates as \textit{Graspabilty}, but ranks them with a prediction network. 

\subsection{Bin-picking Performance}

    First, we evaluate the success rates and time costs for bin-picking experiments (Table \ref{tab:compare}). The number after slash denotes to the total picking times of one trial, and the one before the slash is the number of times when the robot picks up only one object. As a baseline, \textit{Graspability} struggles in success rates since it can not discriminate whether the target is entangled with others or not. Our method and \textit{CNN} both reach relatively higher success rates for picking a single object. Particularly, our method improves the performance of picking from S-shaped objects by a success rate of 50$\%$. 
    The reason why \textit{CNN} struggles with an S-shaped object is that it uses quarters of depth map to make predictions. Even if the cropped image contains the complete shape of target objects, it still lacks information of entangling with others. Our method directly evaluates entanglement for a complete depth map to solve the problem bothering \textit{CNN}. For the mixed objects, our method reaches a high success rate of 70$\%$ since our model-free method only focuses on the information of edges in the depth map. The superior performance of our method indicates that our hand-engineered approach can analyze the relationship between these objects directly and efficiently. \textit{CNN} may require more evaluation for generalization while our method can be utilized without training. 

\subsection{Qualitative Analysis}

    From the examples presented in Fig. \ref{fig:res}, we validate how our method selects graspable objects qualitatively. For the same depth map as input, we use baselines and our method to detect optimal grasp positions, and the top-ranked grasp is marked using green dots. It is observed that our method can directly find the objects that are not tangled with others in the bin, while \textit{Graspability} and \textit{CNN} always focus on objects at the top of the clutter. Especially in the fifth column when all five grasp candidates are classified as tangled, both existing approaches predict poorly while our method successfully finds the graspable objects without any entanglement in the bin. Graspable objects selected by the proposed method are similar to the human observations. The reason is that our method uses edge and topological knowledge to explain the entanglement relationship more intuitively, which guarantees a complete observation of all potential entanglement in the bin. 
    
\begin{table}[t]
\renewcommand\arraystretch{1.2}
\centering
\caption{Success rates of picking one single object}
    \begin{tabular}{@{\extracolsep}clcccc}
    \toprule
    \multicolumn{1}{l}{} & & Graspability\cite{domae2014fast} & CNN\cite{matsumura2019learning} & Ours \\ 
    \midrule
    \multirow{4}{*}{\begin{tabular}[c]{@{}c@{}} Success rate
    \end{tabular}} & C-shaped object & 11/20 & 14/20 & 15/20 \\
    & \begin{tabular}[c]{@{}l@{}} S-shaped object\end{tabular} & 6/20 & 8/20 & 10/20 \\ 
    & \begin{tabular}[c]{@{}l@{}} Mixed objects\end{tabular} & 8/20 & 10/20 & 14/20\\ 
    \cmidrule{2-5}
    \multirow{5}{*}{} & Total & 25/60 & 32/60 & 39/60 \\
    \midrule
    \multirow{1}{*} Time cost (s) &  & 2.1 & 2.7 & 7.8\\
    \bottomrule
    \end{tabular}
\label{tab:compare}
\end{table} 
    
    In addition, the average time costs of \textit{Graspability}, \textit{CNN}, and our method are 2.1s, 2.7s, 7.8s. The time cost of our method depends on how many line segments are detected from the depth map. For our experiment setting of 10 objects placed in the bin, the time cost per trial is limited to 8s. 

\subsection{Discussion}

\begin{figure}[h] 
    \centering
    \includegraphics[width=0.75\linewidth]{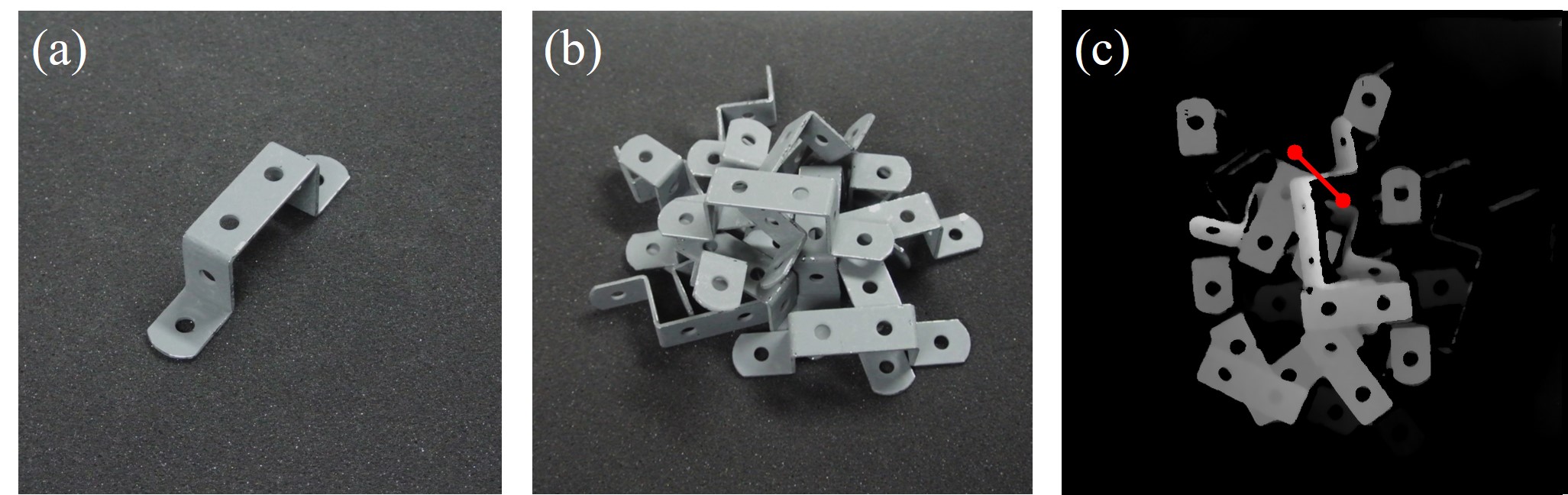} 
    \caption{(a) Our method can not apply to this type of object. (b-c) Piled clutter provides too much information. It makes the topology coordinates can not handle extra edge segments, which leads to fail prediction. }
    \label{fig:exception} 
\end{figure}

    Let us consider other kinds of industrial parts like Fig. \ref{fig:exception} shows. This type of object provides too much edge information for topology coordinates, which may cause some misunderstandings. In this case, a learning-based approach would be necessary. However, since our method prefers objects with the shape of pure edges such as rigid linear objects with a smooth edge, it is possible to develop our method on manipulation of deformable linear objects such as tubes or ropes in future work. 
    
    Common failures that result from our method are caused by the situation where the selected region does not contain any graspable positions. Even if the entanglement map can be generated correctly, grasps can not be detected due to the collisions in the selected region. In the future, it may be possible to add more collision information during generating the entanglement map to improve the performance. 
    
\section{CONCLUSIONS}\label{sec:con}

    In this paper, we present a topology-based solution for a robot to only pick only one object in robotic bin-picking. We present a topological feature map called entanglement map to describe the entanglement situation of cluttered objects in a bin. A grasp synthesis method is proposed to search for the optimal grasp without picking up entangled objects from the whole input image. We reach fine success rates on real-world experiments. Our method is dependable upon its generalization capability even if for mixed objects in bin-picking. Particularly, we do not need large training data to make predictions since the proposed method can obtain the topological relationship of entangled objects even for complex-shaped objects. 

\bibliographystyle{ieeetr}
\bibliography{ebibsample.bib}
\end{document}